\documentclass{article}

\usepackage{arxiv}

\usepackage[utf8]{inputenc} 
\usepackage[T1]{fontenc}    
\usepackage{hyperref}       
\usepackage{url}            
\usepackage{booktabs}       
\usepackage{amsfonts}       
\usepackage{nicefrac}       
\usepackage{microtype}      

\usepackage{graphicx}
\usepackage{pstricks-add}
\usepackage{multirow}
\usepackage[autostyle]{csquotes}
\usepackage{siunitx}

\title{Leveraging the Self-Transition Probability of Ordinal Pattern Transition Graph for Transportation Mode Classification}

\author{
  Isadora Cardoso-Pereira\thanks{This work is the english version of \url{https://doi.org/10.5753/sbrc.2019.7391}} \\
  Federal University of Minas Gerais\\
  Belo Horizonte, Brazil \\
  \texttt{isadoracardoso@dcc.ufmg.br} \\
  \And
  Jo\~ao B. Borges \\
  Federal University of  Rio \\ Grande do Norte\\
  Natal, Brazil \\
  \texttt{joaoborges@dct.ufrn.br}\\
  \And
  Pedro H. Barros \\
  Federal University of Minas Gerais \\
  Belo Horizonte, Brazil \\
  \texttt{pedro.barros@dcc.ufmg.br} \\
  \And
  Antonio F. Loureiro \\
  Federal University of Minas Gerais \\
  Belo Horizonte, Brazil \\
  \texttt{loureiro@dcc.ufmg.br} \\
  \And
  Osvaldo A. Rosso \\
  Federal University of Alagoas \\
  Macei\'o, Brazil \\
  \texttt{oarosso@gmail.com} \\
  \And
  Heitor S. Ramos \\
  Federal University of Minas Gerais \\
  Belo Horizonte, Brazil \\
  \texttt{ramosh@dcc.ufmg.br} 
}

\begin{document}
\maketitle

\begin{abstract}
    The analysis of GPS trajectories is a well-studied problem in Urban Computing and has been used to track people. 
    Analyzing people mobility and identifying the transportation mode used by them is essential for cities that want to reduce traffic jams and travel time between their points, thus helping to improve the quality of life of citizens. The trajectory data of a moving object is represented by a discrete collection of points through time, i.e., a time series. 
    Regarding its interdisciplinary and broad scope of real-world applications, it is evident the need of extracting knowledge from time series data. 
    Mining this type of data, however, faces several complexities due to its unique properties. 
    Different representations of data may overcome this. 
    In this work, we propose the use of a feature retained from the Ordinal Pattern Transition Graph, called the probability of self-transition for transportation mode classification.
    The proposed feature presents better accuracy results than Permutation Entropy and Statistical Complexity, even when these two are combined. 
    This is the first work, to the best of our knowledge, that uses Information Theory quantifiers to transportation mode classification, showing that it is a feasible approach to this kind of problem.
\end{abstract}


\section{Introduction}
The analysis of GPS trajectories is a well-studied problem in Urban Computing and has been used to track people~\cite{understanding-mobility-based-on-gps-data}, vehicles~\cite{DBLP:journals/corr/AquinoCAFR15}, animals~\cite{handcock2009monitoring},  and meteorological events~\cite{wang2015long}. 
In particular, the analysis of people's mobility and the identification of their transportation mode are essential activities for cities that want to reduce traffic jam and travel time, thus helping to improve the life quality of their citizens. 
A discrete collection of points represents the trajectory data of a moving object through time, i.e., a time series. 
There is a wide range of fields that study their phenomena using temporal observations, such as astrophysics (e.g., solar radiation~\cite{REIKARD2009342}), medicine (e.g., cardiac diseases~\cite{doi:10.1117/12.2254704}), among many others.

Regarding its interdisciplinary and broad scope of real-world applications, it is evident the need of extracting knowledge from time series data. 
Therefore, they have been the subject of study for decades~\cite{LANGKVIST201411}. Mining this type of data, however, faces several complexities and is considered by Yang and Wu~\cite{yang200610} one of the most challenging problems in data mining research due to its unique properties. 
Besides high dimensionality, heterogeneity and noise, well-known problems in the big data era, time series depend on the ordering, and, thus, a change in the order could change their meaning. 
It opposes the common assumption made by many algorithms, such as Na\"ive Bayes, of independent and identically distributed observations, leading standard classification methods to perform poorly in time series~\cite{DBLP:journals/corr/BagnallBLL16}. 
Different representations of data may overcome this, as discussed in the following.

Data representation, or data pre-processing, is an essential step in time series data mining. It consists of applying a transformation directly to the time series into the same time domain, such as summarizing original data points into more comprehensible format~\cite{wilson2017data}, or change the data from the time domain to another domain, e.g., frequency, shapelets, symbol-based~\cite{DBLP:journals/corr/BagnallBLL16}. 
A proper representation should not only reduce the dimensionality and remove random noise, but it must preserve the critical local and global features of the original data as well~\cite{wilson2017data}. 
Hence, useful features, i.e., features that represent the original data, can be used to classify time series data, enabling an efficient computation. 
Also, these features should be robust to data problems, such as data missing, outliers, irregular time spacing, for instance. 

In this context, the problem addressed in this work is:

\begin{displayquote}
\textit{Given a time series of consecutive localization, is it possible to obtain useful features capable of characterizing, and, therefore, classify such time series in terms of the transportation mode used by the user?}
\end{displayquote}

This analysis is based on Information Theory methods, such as Ordinal Patterns (OP)~\cite{PhysRevLett.88.174102}. 
OP is a model-free method based on the sequence that naturally arises from the time series, comparing the values that are in the same neighborhood and replacing them with a sequence of symbols. 
Along with OP, we use its graph transformation, known as Ordinal Pattern Transition Graph (OPGT)~\cite{small2013complex}, to represent the time series data in a new domain, and, then, classify it using features taken from such transformation. 
Therefore, we propose the use of a new feature, derived from OPGT, called Probability of Self-Transition ($p_{ST}$). 

Having these tools, we aim to characterize the time series according to their behavior. 
The validation of our proposal is made in a real-world problem of Urban Computing, referring to transportation mode classification: we want to characterize which transportation mode (car, bus, bike, and walk) a given person carrying a GPS is traveling. 
The contribution of this work is twofold:
\begin{itemize}
    \item the proposal of a new feature to characterize and classify time series;
    \item the use of Information Theory methods to transport mode classification.
\end{itemize}{}
These are important contributions to advance the state of the art of mobility analysis.

This work is organized as follows. 
Section~\ref{sec:related} discusses the related work.
Section~\ref{sec:methodology} presents the methodology used in this work.
Section~\ref{sec:results} discusses the obtained results.
Finally, Section~\ref{sec:conclusion} concludes this work.

\section{Related Work} \label{sec:related}

The characterization and classification of time series is the subject of study of several areas and, as such, is widely explored. 
There are several contributions in the field of Machine Learning (ML), as can be seen in~\cite{DBLP:journals/corr/abs-1012-2789, DBLP:journals/corr/BagnallBLL16, Lines:2018:TSC:3234931.3182382}. 
These studies compare the effectiveness of more than $20$ ML techniques in the classification of time series of diverse domains, including Urban Computing (e.g., pedestrian and car counting to understand the use of public spaces, prediction of events such as earthquakes from sensors deployed throughout the city, etc). 
Among the evaluated techniques, Hierarchical Vote Collective of Transformation-based Ensembles (HIVE-COTE) stands out. 
Proposed by Lines et {al.}~\cite{Lines:2018:TSC:3234931.3182382}, HIVE-COTE is an ensemble technique composed of $35$ classifiers, modularized according to the domain in which they act. 
Although it presents a good overall accuracy in several domains, it is a technique of high computational cost, even in comparison to Deep Learning (as seen in~\cite{fawaz2018deep}), which makes it unfeasible in high dimensional time series.

Techniques derived from Information Theory have also been successful in the characterization of time series in the area of Urban Computing. 
Such methods can distinguish time series using model-free techniques that also are computationally inexpensive and have low dimensionality, such as OP and Complexity-Entropy Plane~\cite{PhysRevLett.99.154102:distinguishing}. 
For instance, Aquino et {al.}~\cite{DBLP:journals/corr/AquinoCAFR15} characterized the behavior of vehicles through their velocities; 
Aquino et {al.}~\cite{AQUINO2017277} characterized the behavior of electric loads, and Ribeiro et {al.}~\cite{PhysRevE.95.062106} characterized the behavior of the crude oil price.

Another research direction that has also been successful in the characterization of time series is based on the transformation of the time series into graphs. 
Using this strategy, networks that inherit the characteristics of the original time series are constructed (for example, periodic series are transformed into regular graphs, and random series are transformed into random graphs). 
Some examples are the visibility graph~\cite{Lacasa4972} and the horizontal visibility graph~\cite{PhysRevE.80.046103}. 
However, as each time series sample is transformed into a vertex of the graph, there is an impact on the scalability of these techniques, making them not feasible for high-dimensional time series.

Recently, methods that combine more than one approach are emerging. 
In~\cite{small2013complex},~\cite{10.1371/journal.pone.0108004},~\cite{zhang2017constructing}, and~\cite{guo2018cross}, we can see techniques that obtain graphs from permutations of possible patterns in OP, taking advantage of the two approaches.

Many proposals study transportation mode classification, as we can see in~\cite{understanding-mobility-based-on-gps-data, dabiri2018inferring, shin2015urban}. However, none of them use an Information Theory approach, as we show in this work can take advantage of this technique.

Studies, as mentioned above, show that time series classification, especially transportation mode classification, is possible. 
This work is inspired by the use of different areas, as well as by their combination. 
Here, we obtain a new feature retained from OPGT and evaluate its impact on the time series classification in the context of Urban Computing.
\section{Methodology} \label{sec:methodology}

\subsection{Dataset} \label{sec:basededados}

In this work, we use the GeoLife\footnote{\url{https://www.microsoft.com/en-us/download/details.aspx?id=52367}} data, collected by Zheng et {al.}~\cite{understanding-mobility-based-on-gps-data}. 
This dataset presents GPS trajectories of $182$ users over five years (from April 2007 to August 2012), containing latitude, longitude, and altitude information.

Among these users, $73$ have transportation mode information, which will be classified in this study. 
Note that only transportation mode with a duration higher than $1000$ hours were considered, since we understand that the smaller the time series, the more difficult to extract relevant information, which leads to the generation of low-quality models.
Table~\ref{tab:transporte} describes the transportation mode used in this work. We have four kinds of transportation: walking, bike, bus, and personal car (car/taxi). 

\begin{table}[htpb]
\caption{Distance and duration of transportation mode}
\label{tab:transporte}
\centering
\begin{tabular}{@{}ccc@{}}
\toprule
\multicolumn{1}{l}{Transport} & \multicolumn{1}{l}{distance (km)} & \multicolumn{1}{l}{duration (h)} \\ \midrule
\multicolumn{1}{l}{walking} & \multicolumn{1}{l}{\num{10123}} & \multicolumn{1}{l}{\num{5460}} \\
\multicolumn{1}{l}{bike} & \multicolumn{1}{l}{\num{6495}} & \multicolumn{1}{l}{\num{2410}} \\
\multicolumn{1}{l}{car/taxi} & \multicolumn{1}{l}{\num{32866}} & \multicolumn{1}{l}{\num{2384}} \\
\multicolumn{1}{l}{bus} & \multicolumn{1}{l}{\num{20281}} & \multicolumn{1}{l}{\num{1507}} \\ \bottomrule
\end{tabular}
\end{table}

Here, we define trajectory as an uninterrupted sequence of GPS points (latitude and longitude) that belong to the same transportation mode. We consider that every user is at the same altitude, thus discarding this measure. 
Also, we discard trajectories with less than $10$ points so we may avoid the creation of low-quality trajectories, which may affect the generated model. 
Table~\ref{tab:trajetorias} shows the total of trajectories obtained from each transportation.

\begin{table}[htpb]
\caption{Obtained trajectories from the dataset}
\label{tab:trajetorias}
\centering
\begin{tabular}{@{}cc@{}}
    \toprule
        \multicolumn{1}{l}{transport} & \multicolumn{1}{l}{trajectories} \\ 
    \midrule
        \multicolumn{1}{l}{walking} & \multicolumn{1}{l}{\num{1653}} \\
        \multicolumn{1}{l}{bike} & \multicolumn{1}{l}{\num{840}} \\
        \multicolumn{1}{l}{bus} & \multicolumn{1}{l}{\num{1017}} \\
        \multicolumn{1}{l}{car/taxi} & \multicolumn{1}{l}{\num{831}} \\
        \addlinespace
        \multicolumn{1}{l}{total} & \multicolumn{1}{l}{\num{4341}} \\ 
    \bottomrule
\end{tabular}
\end{table}

\subsection{Ordinal Pattern Transformation}

Ordinal Patterns (OP) is a simple method of transforming time series that does not require any model assumption about the time series and can be applied to any arbitrary time series. 
Furthermore, such a method has an advantage of its simplicity, speed, robustness, and invariance concerning non-linear monotonic transformations. 
This approach is based on the sequence that naturally arises from the time series, comparing the values that are in the same neighborhood and replacing them with a sequence of symbols~\cite{PhysRevLett.88.174102}.

Let a temporal series $\mathbf{X}(t) = \{x_1, x_2, {\ldots}, x_n \}$ of size $n$ and let also an embedding dimension $D \in \mathbb{N}$ and an embedding delay $\tau \in \mathbb{N}$. 
In each time instant $t = \{1, {\ldots}, n - (D - 1) \tau\}$, we have a sliding window $w_t \subseteq x$, such as
$$w_t = \{x_t, x_{t + \tau}, {\ldots}, x_{t + (D - 2)\tau},  x_{t + (D - 1)\tau} \},$$
i.e., each element within the sliding window is obtained from the time series in the time $t, {\ldots}, t + (D - 1) \tau$. 
This corresponds to a time series sample at evenly spaced intervals.

The ordinal relation for each instant $t$ consists of the permutation $\pi~=~\{r_0, r_1, {\ldots}, r_{D-1}\}$ of $\{0, 1, {\ldots}, D - 1\}$, so that

$$x_{t-r_{D-1}} \leq x_{t-r_{D-2}} \leq \cdots \leq x_{t-r_{1}} \leq x_{t-r_0}.$$

In other words, $\pi$ represents the permutation of elements in the sliding window $w_t$, in ascending order. 
In order to obtain unique results, we define that, if a time series have elements such that $x_{t-r_{i}} = x_{t-r_{i - 1}}$, we consider that $r_i < r_{i-1}$.
Hence, the time series is converted to a set of ordinal patterns, $\Pi = \{ \pi_1, {\ldots}, \pi_m \}$, where $m = n - (D - 1)\tau$ and each $\pi_m$ represents a permutation of the possible permutation set of $D!$~\cite{AQUINO2017277}.

The choosing of $D$ depends on the time series size and must satisfy the condition $n \gg D!$ -- the higher $D$ is, the greater the time series length  is necessary to have reliably extracted data~\cite{PhysRevLett.99.154102:distinguishing}. 
If interested, more explanations are given in~\cite{staniek2007parameter}. 
For practical purposes, Bandt and Pompe~\cite{PhysRevLett.88.174102} recommend values such that $3 \leq D \leq 7$, which are adopted in this work.

For all $D!$ possible permutation $\pi$ of $D$, the relative frequency can be computed by the times a certain sequence appeared in the time series, divided by the number of total sequences, obtaining the histogram of the probability distribution $P \equiv \{p(\pi)\}$, which is defined by:
$$p(\pi) = \frac{\mid s_\pi \mid}{n - (D - 1) \tau }, $$
where $\mid s_\pi \mid \in \{ 0, {\ldots}, m \}$ is the number of pattern observed of type $\pi$.

From this new representation, it is possible to extract features, such as Information Theory quantification, which can be used to characterize the time series dynamics~\cite{PhysRevLett.99.154102:distinguishing}. 
In this work, we extract two quantifiers, the Permutation Entropy, and Statistical Complexity, as discussed in the following.

\subsubsection{Permutation Entropy}

The Permutation Entropy is a measure of uncertainty associated with the process described by $p_\pi$ and is defined by:
$$H[p_\pi] = - \sum p(\pi) \log_2 p(\pi),$$
where $0 \leq H[p_\pi] \leq log D!$. 
This measure is equivalent to the Shannon Entropy~\cite{AQUINO2017277}. 
Low values of $H[p_\pi]$ represent a sequence of increasing or decreasing values in the permutation distribution, indicating that the original time series is deterministic, while high values indicate a completely random system~\cite{PhysRevLett.88.174102}.

The maximum value for $H[p_\pi]$ occurs when all possible permutations of $D!$ have the same probability of occurring, which is the case for the uniform distribution $p_u$ of permutation. 
Thus, $H_{max} = H[p_u] = \log D!$~\cite{PhysRevE.86.046210}. 
We can define the normalized Shannon Entropy, for the case of permutation entropy, as:
\begin{equation}
    H_S[p_\pi] = \frac{H[p_\pi]}{H_{max}},
    \label{eq:shannon}
\end{equation}
where $0 \leq H_S[p_\pi] \leq 1$~\cite{PhysRevLett.99.154102:distinguishing}.

\subsubsection{Statistical Complexity}

The Statistical Complexity is based on Jensen-Shannon divergence (JS) between the associated probability distribution $p_\pi$ and the uniform distribution $p_u$ (the trivial case for the minimum knowledge of the process) and is defined by:
$$C_{JS} [p_\pi] = Q_{JS}[p_\pi,p_u] H_S [p_\pi], $$
where $p_\pi = \{ p(\pi) \}$ is the probability distribution of ordinal patterns, $p_u$ is the uniform distribution and $H_S$ is the normalized Shannon Entropy, as defined in Equation~\ref{eq:shannon}. 
The disequilibrium $Q_{JS}[p_\pi,p_u]$ is given by:
$$ Q_{JS}[p_\pi,p_u] = Q_{0} JS[p_\pi,p_u] = Q_0 \left(S \left[\frac{p_\pi + p_u}{2}\right] - \frac{S[p_\pi] + S[p_u]}{2}  \right),$$
where $S$ is the Shannon entropy and $Q_0$ is defined by:
$$ Q_0 = -2 \left[ \left( \frac{D! + 1}{D!} \right) ln(D! + 1) - 2ln(2D!) + ln(D!)  \right]^{-1},$$
which describes the normalization constant, which is equal to the inverse of the maximum value of $JS[p_\pi, p_u]$ e $0 \leq Q_{JS} \leq 1$~\cite{AQUINO2017277, PhysRevLett.99.154102:distinguishing}.

\subsection{Ordinal Pattern Transition Graph}

Given a sequence of OP $\Pi$, the OPTG represents the relation between consecutive patterns and is defined as a weighted directed graph $G_\pi = (V, E)$, with vertices $v_{\pi_i} \in V = \{ v_{\pi_i} : i = 1, {\ldots}, D!\}$ that correspond to a possible permutation of $D!$ to the embedding dimension $D$, and edges $E = \{ (v_{\pi_i}, v_{\pi_j}): v_{\pi_i}, v_{\pi_j} \in V\}$.

A directed edge connects two OPs in the graph if such patterns appear sequentially in the original time series, representing a transition between the patterns. 
The weights $w: E \rightarrow \mathbb{R}$ of the edges represent the probability of existence of a specific transition in $\Pi$ and is given by:
$$w(v_{\pi_i}, v_{\pi_j}) = \frac{\mid \Pi_{\pi_i, \pi_j} \mid }{m - 1}, $$
where $\mid \Pi_{\pi_i, \pi_j} \mid \in \{0, {\ldots}, m - 1 \}$ is the number of transitions between the permutations $\pi_i$ e $\pi_j$ and $\sum_{v_{\pi_i}, v_{\pi_j}} w(v_{\pi_i}, v_{\pi_j}) = 1$. 

Once the graph is constructed from the OP set, some properties are inherited from this transformation. The most notable are:
\begin{itemize}
    \item \textbf{simplicity and speed}: the graph construction only depends on the number $m$ of OPs, needing to count the number of transitions in $m - 1$ steps. 
    In turn, the time series transformation into OP depends on the size $n$ of the time series and the embedding dimension $D$. 
    The complexity of this transformation is limited by $O(nD^2)$, assuming that the permutation is obtained by ordering the sliding windows by a simple sorting algorithm, such as Selection Sort, in $O(D^2)$ and $\tau = 1$, in the worst case. 
    For practical reasons, since $D$ is recommended to be in the interval between \num{3} and \num{7}, the ordering of this strategy has a maximum of \num{7} elements, so the complexity of such strategy is more dependent on time series size $n$;
    \item \textbf{scalability}: the approaches that use a visibility graph~\cite{Lacasa4972}, for instance, transform each time series sample into a vertex within the graph -- an impracticable approach to high-dimensional time series due to the space required for storage. 
    On the other hand, the number of vertices of the OPGT is given by the embedding dimension $D$, not depending on the size of the series and being limited by $D!$.
    \item \textbf{robustness}: OPs are robust to the presence of noise and invariants with respect to non-linear monotonic transformations~\cite{AQUINO2017277, PhysRevLett.99.154102:distinguishing}.
\end{itemize}

\begin{figure}[htpb]
    \centering
    \includegraphics[width=0.7\linewidth]{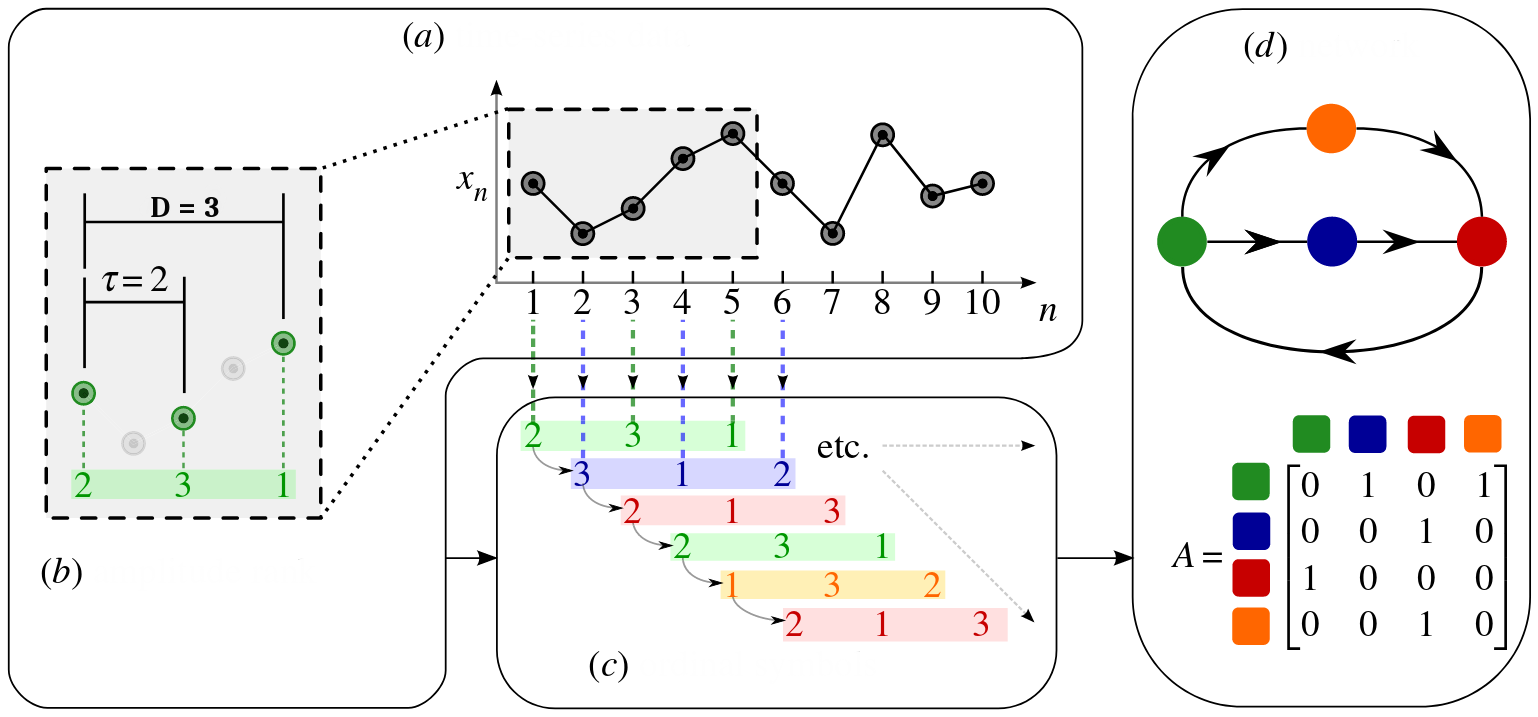}
    \caption{Illustration of time series transformation to OPTG (Adapted from~\cite{mccullough2017multiscale})}
    \label{fig:ilustracao}
\end{figure}

Figure~\ref{fig:ilustracao} illustrates the process described above:
(a) Given a time series; 
(b) we calculate the sliding windows with values for $D$ and $\tau$ ($D = 3$ and $\tau = 2$, in the figure); 
(c) we obtain the OP; and
(d) we build the OPTG with a vertex to each OP found in the time series and with edges that describe the temporal succession of patterns~\cite{mccullough2017multiscale}.

\subsection{Probability of Self-Transition}

The self-transitions of the transition graph are the edges from a vertex to itself, also known as loop. Its presence in a graph represents the occurrence of the same OP consecutively. 

Zhan et {al.}~\cite{zhang2017constructing} proposed an analysis of the entropy computed through the weights of the edges of the transition graph after the removal of the self-transition edges. 
However, these transitions are directly related to the temporal correlation of the original time series and are a valuable indication of the hidden dynamics and, therefore, should not be discarded. 
The way these edges are placed is an essential element for the subsequent analysis of the graph.

The probability of self transition is defined as the probability of occurrence of a sequence of equal patterns within the OP set and can be expressed as:
$$p_{st} = p(\pi_i, \pi_i) = \sum_{i \in 1, {\ldots}, D!} w(v_{\pi_i}, v_{\pi_i}).$$

The weight normalization of the graph adopted in this work is similar to that adopted by Zhan et {al.}~\cite{zhang2017constructing}, where the authors normalize the weights such that all weights sum $1$. 
However, in our case, we accept the presence of self-transition.

\section{Results} \label{sec:results}

We evaluate the quality of our proposal through transportation mode classification, using the dataset described in Section~\ref{sec:basededados}. 
In this case, we assume that a good quality refers to good accuracy, precision, and sensitivity results in the classification -- presenting good results in such metrics, we infer that the data representation used in this work is satisfactory, i.e., the proposed feature retains information of the original time series, making possible its classification.

\begin{figure}[htpb]
\centering
\includegraphics[width=0.7\linewidth]{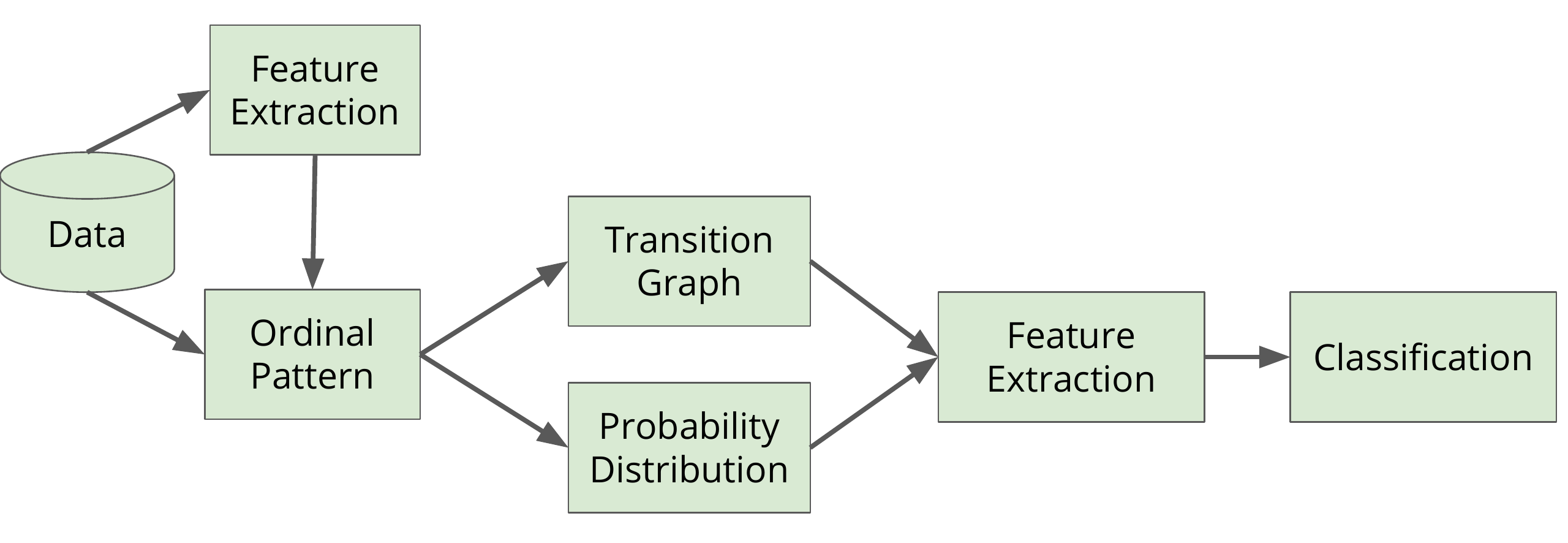}
\caption{Methodology process applied in this work}
\label{fig:metodologia}
\end{figure}

The process used to extract features and afterward perform the classification is shown in Figure~\ref{fig:metodologia}. 
Depending on the nature of the data, it is possible to use them without extracting any previous information, as well as extracting features that can better express the hidden knowledge of the data, highlighting them in the transformation. 
In this work, we use the two approaches for comparison: we use the latitude and longitude provided by the data, and we also transform these two attributes into a third one, the Euclidean distance between two consecutive points. 
Then, we transform these three features into PO, from where we build the OPTG and the probability distribution of OP. 
From the OPTG, we extract the probability of self-transition, and from the probability distribution, we extract the Permutation Entropy and Statistical Complexity. 
With these features, we perform the classification.

Since the idea of this work is to highlight the particularity of each data transformation, little effort was devoted to the adjustment of the classification algorithms. 
It may be possible to obtain better results of the evaluation metrics by adjusting the parameters of the classifier. However, our objective is not only to present good results of such metrics but to know if our proposal is suitable for characterization and classification of time series. 
Our classification was made using simple algorithms, which are: 
k-Nearest Neighbors (k-NN), with $ k = 2 $; 
Support Vector Machines (SVM), with linear (SVM-L) and radial (SVM-R) kernels; 
and Decision Tree.

To evaluate if our proposal is capable of generalizing, and also to validate our results, we use cross-validation, with $5$-folds. 
It is important to note that this cross-validation is performed in the extracted features, immediately prior to classification. 
Such features, after going through the transformations described in this work, can be interpreted as independent and identically distributed, allowing the use of this validation method.

The methods used here were implemented in R (version 3.4.4) on a machine with the following configuration: Linux OS, 8 GB RAM and Intel\textregistered ~ Core\texttrademark ~ i5-2410M CPU @ 2.30GHz.

First, we will evaluate the influence of the embedding dimension of $D$ in the classification. Figure~\ref{fig:acuracia_d} shows the obtained accuracy in the classification of latitude, longitude and distance, using different values of $D$ (from \num{3} to \num{7}, inclusive, as recommended in~\cite{PhysRevLett.88.174102}). 
When classifying each feature in isolation, we see that $p_{st}$ achieves the best accuracy value, about \SI{70}{\percent}, while $H[p_\pi]$ and $C_{JS}[p_\pi]$ achieve about \SI{65}{\percent} and \SI{63}{\percent}, respectively. 
That is, using only a single feature, $p_{st}$, there is a gain of about \SI{5}{\percent} of accuracy. 
In Figure~\ref{fig:acuracia_d}, we also see the classification using a set of features: 
\textbf{(1)} $\{H[p_\pi], C_JS[p_\pi]\}$ and \textbf{(2)} $\{H[p_\pi], C_JS[p_\pi], p_{st}\}$. 
For the first set, $p_{st}$ still yields significantly higher results, which suggests that the information gain for $p_{st}$ is greater than the gain for the other two features together. 
In the second set, we see that this classification presents the best results among those presented, that is, the classification using $p_{st}$ can be improved if combined with other features. 
Besides, we see that the best-performing classifier is SVM-R and $D = 5$ is the best value for $D$ in this case.

\definecolor{chocolate(web)}{rgb}{0.82, 0.41, 0.12}
\definecolor{chartreuse(web)}{rgb}{0.5, 1.0, 0.0}
\definecolor{cornflowerblue}{rgb}{0.39, 0.58, 0.93}
\definecolor{coralred}{rgb}{1.0, 0.25, 0.25}

\begin{figure}[htpb]
\centering
\includegraphics[width=0.8\linewidth]{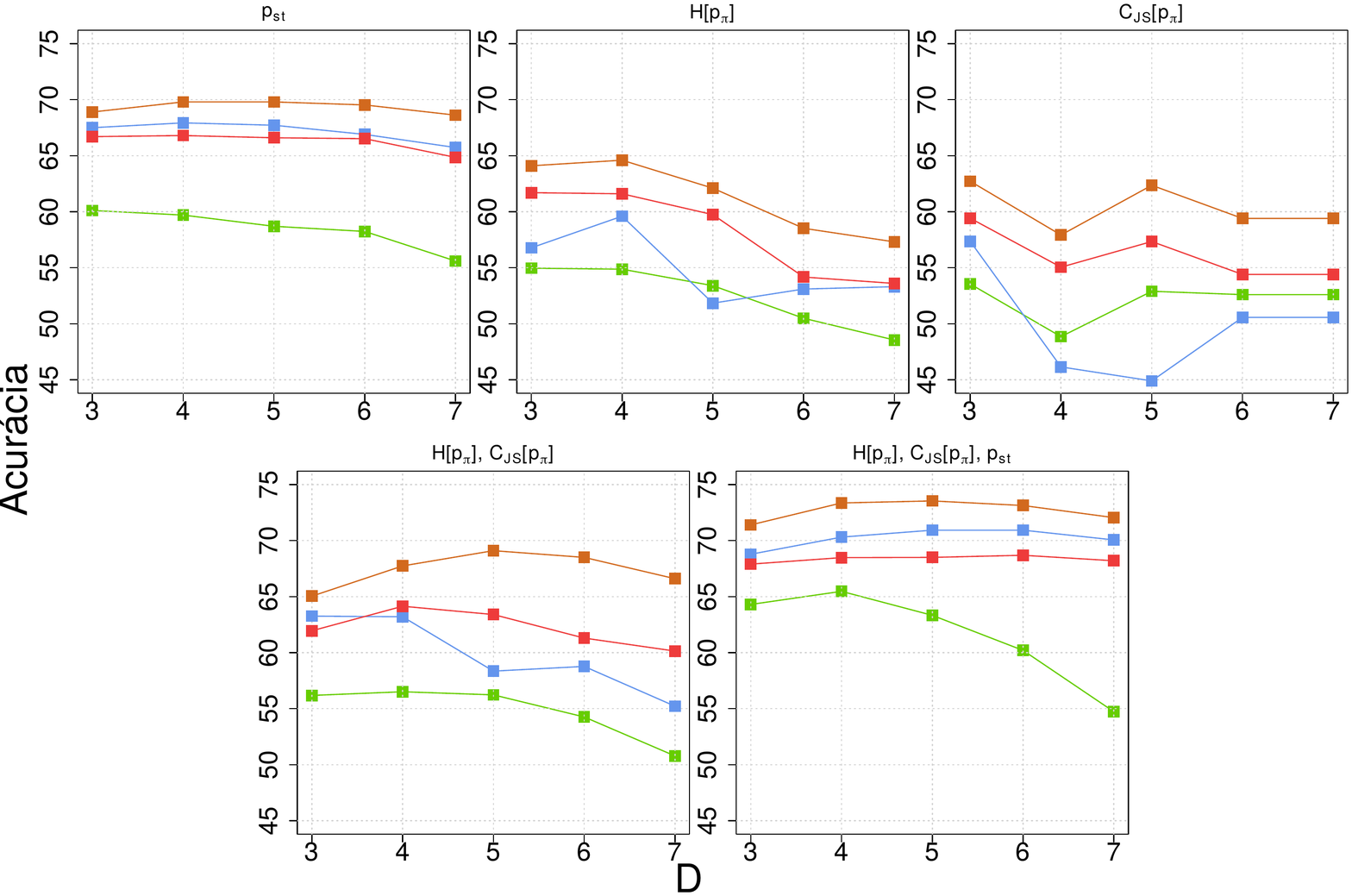}
\caption{Classification Accuracy using SVM-R~({\color{chocolate(web)} \rule[1.25 pt]{4 pt}{4 pt}}), 
SVM-L~({\color{cornflowerblue} \rule[1.25 pt]{4 pt}{4 pt}}), 
DT~({\color{coralred} \rule[1.25 pt]{4 pt}{4 pt}}) and
k-NN~({\color{chartreuse(web)} \rule[1.25 pt]{4 pt}{4 pt}}), for different values of $D$. 
}
\label{fig:acuracia_d}
\end{figure}

Now, we will evaluate the $\tau$ influence in the classification. The maximum $\tau$ value depends on the time series size $n$ and the dimension $D$, being limited by:
$$ \tau < \frac{n}{D - 1}.$$

The greater the $\tau$ value, the greater the number of time series samples. For example, for $D = 5$ and $\tau = 1$, the time series has to be, at least, greater than $4$; 
for $D = 5$ and $\tau = 2$, $n > 8$; 
for $D = 5$ and $\tau = 3$, $n > 12$; 
$D = 5$ and $\tau = 5$, $n > 20$; 
$D = 5$ and $\tau = 10$, $n > 40$; 
$D = 5$ and $\tau = 15$, $n > 60$, and so on.

In other words, as $\tau$ increases, the longer the trajectories must be. 
With this, smaller trajectories, and, consequently, data, are discarded. 
Table~\ref{tab:trajetorias_geral} shows how the $\tau$ value impacts the time series size $n$.

\begin{table}[htpb]
\caption{Values of $n$ as $\tau$ increases}
\label{tab:trajetorias_geral}
\centering
\begin{tabular}{@{}ccccccc@{}}
\toprule
\textbf{$\tau$} & 1 & 2 & 3 & 5 & 10 & 15 \\ \midrule
\textbf{$n$} & 4341 & 4329 & 4290 & 4201 & 4024 & 3871 \\ \bottomrule
\end{tabular}
\end{table}

\begin{figure}[htpb]
\centering
\includegraphics[width=0.8\linewidth]{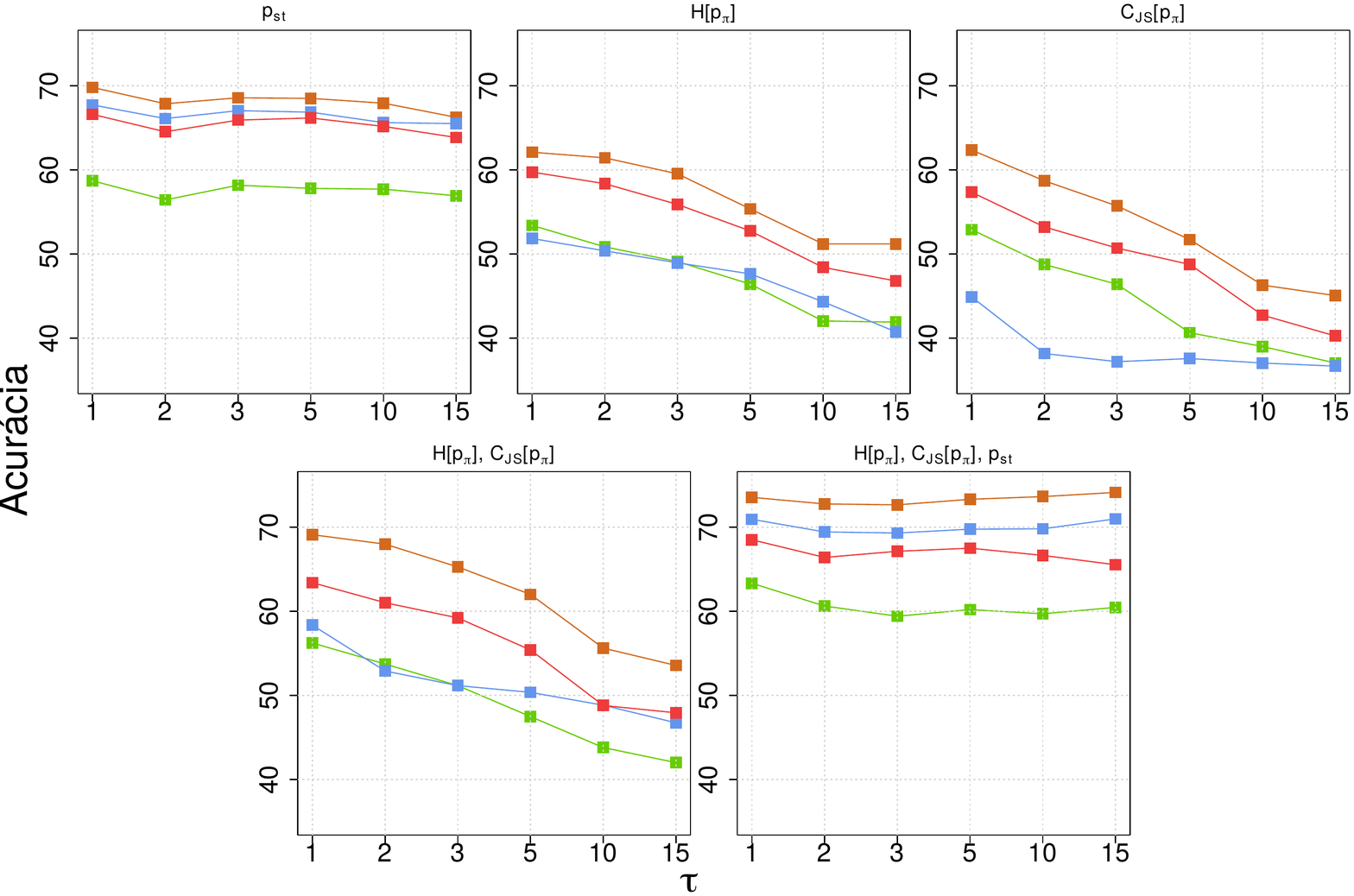}
\caption{Classification Accuracy using SVM-R~({\color{chocolate(web)} \rule[1.25 pt]{4 pt}{4 pt}}), 
SVM-L~({\color{cornflowerblue} \rule[1.25 pt]{4 pt}{4 pt}}), 
DT~({\color{coralred} \rule[1.25 pt]{4 pt}{4 pt}}) and 
k-NN~({\color{chartreuse(web)} \rule[1.25 pt]{4 pt}{4 pt}}), for different values of $\tau$. 
}
\label{fig:acuracia_tau}
\end{figure}

In Figure~\ref{fig:acuracia_tau}, we see the classification for $D = 5$ (the $D$ value that presented the best accuracies, shown in Figure~\ref{fig:acuracia_d}) and different values of $\tau \in \{1,2,3,5,10,15\}$. 
It is possible to note that, among the features classified in isolation, $p_{st}$ presents the most stable behavior, suffering less variation of accuracy as $\tau$ varies. 
On the other hand, $H[p_\pi]$ and $C_{JS}[p_\pi]$ presents accuracy values that decrease as $\tau$ increases, both when classified alone and together. 
It is also possible to note that $p_{st}$ conserves its robustness in relation to $\tau$ values even when classified with the other features. 

From the presented results, we can understand that, even with minor time series, there is no abrupt compromise of the yield of $p_{st}$. 
In other words, $p_{st}$ has less dependence on $D$ and $\tau$ values. 

Moreover, in Figure~\ref{fig:acuracia_tau} we can also note that $\tau = 1$, in general, presents better results in all the classified sets. This makes sense because as the $\tau$ value increases, more information may be lost within the trajectory due to the distance between the points in the sliding window.

Furthermore, we have SVM-R again as the classifier with the best accuracy values.

After that, we will adopt the values $D = 5$ and $\tau = 1$ and the SVM-R classifier to analyze in more details the time series classes of the used dataset. 
The classification was done using the three features $p_{st}$, $H[p_\pi]$ and $C_{JS}[p_\pi]$, since, as we saw earlier, this set obtains a better accuracy value. 

Table~\ref{tab:conj_acc} shows the classification for the set of classes presented in the dataset, along with the confidence interval (with 95\% confidence). 
Besides accuracy, we used sensitivity (sen), precision (pre), and F1-score as evaluation metrics, defined as:

\begin{itemize}
    \item 
        The sensibility (sen) explains how effectively the classifier identifies positive predictions. 
        That is, the ability of our model to identify  which individuals pertain to a class;
    \item 
        The precision (pre) express the proportion of points in data which the model says that they are relevant and they are;
    \item 
        F1-score is the harmonic mean between precision and sensibility.
\end{itemize}

It is possible to see that there are more challenging to distinguish between transportation that, intuitively, travels at a similar pace, as walking and bike, and car/taxi and bus. 
For more distinct transport, best results are achieved.

\begin{table}[htpb]
\centering
\caption{Evaluation metrics of classification in used dataset}
\label{tab:conj_acc}
\scalebox{0.85}{
    \begin{tabular}{clllc}
    \hline
    classes & \multicolumn{1}{c}{Pre} & \multicolumn{1}{c}{Sen} & \multicolumn{1}{c}{F1} & Accuracy \\ \hline
    \addlinespace
        walking & 93,98\% ($\pm 1,42$) & 80,66\% ($\pm 1,05$) & 86,77\% ($\pm 0,93$) & \\
        bike & 55,83\% ($\pm 3,57$) & 82,46\% ($\pm 4,86$) & 66,45\% ($\pm 3,85$) & \multirow{-2}{*}{81,05\% ($\pm 1,52$)} \\
    \addlinespace
        walking & 96,47\% ($\pm 0,66$) & 91,20\% ($\pm 0,77$) & 93,75\% ($\pm 0,65$) &  \\
        bus & 84,75\% ($\pm 1,67$) & 93,69\% ($\pm 1,32$) & 88,96\% ($\pm 1,46$) & \multirow{-2}{*}{92,03\% ($\pm 0,90$)} \\
    \addlinespace
        walking & 97,46\% ($\pm 0,81$) & 88,62\% ($\pm 1,22$) & 92,82\% ($\pm 0,50$) & \\
        car/taxi & 75,20\% ($\pm 2,75$) & 93,70\% ($\pm 1,75$) & 83,40\% ($\pm 1,41$) & \multirow{-2}{*}{90,00\% ($\pm 0,71$)} \\
    \addlinespace
        bike & 93,40\% ($\pm 1,43$) & 87,40\% ($\pm 2,37$) & 90,26\% ($\pm 1,64$) &  \\
        bus & 88,81\% ($\pm 1,80$) & 94,10\% ($\pm 0,80$) & 91,37\% ($\pm 1,00$) & \multirow{-2}{*}{90,86\% ($\pm 1,21$)} \\
    \addlinespace
        bike & 92,40\% ($\pm 2,00$) & 86,86\% ($\pm 1,20$) & 89,53\% ($\pm 1,15$) &  \\
        car/taxi & 85,85\% ($\pm 1,24$) & 91,70\% ($\pm 1,94$) & 88,67\% ($\pm 0,85$) & \multirow{-2}{*}{89,13\% ($\pm 1,01$)} \\
    \addlinespace
        bus & 81,05\% ($\pm 3,82$) & 85,12\% ($\pm 0,95$) & 83,00\% ($\pm 2,04$) &  \\
        \multicolumn{1}{l}{car/taxi} & 82,60\% ($\pm 1,90$) & 78,11\% ($\pm 2,85$) & 80,24\% ($\pm 1,33$) & \multirow{-2}{*}{81,74\% ($\pm 1,60$)} \\
    \addlinespace
        walking & 90,53\% ($\pm 2,03$) & 68,25\% ($\pm 1,82$) & 77,74\% ($\pm 1,10$) & \\
        bike & 51,16\% ($\pm 2,53$) & 78,22\% ($\pm 4,93$) & 61,76\% ($\pm 3,05$) & \\
        bus & 70,75\% ($\pm 2,03$) & 81,06\% ($\pm 1,82$) & 75,40\% ($\pm 1,10$) & \\
        car/taxi & 65,93\% ($\pm 4,15$) & 77,13\% ($\pm 2,32$) & 70,95\% ($\pm 2,32$) & \multirow{-4}{*}{73,54\% ($\pm 0,70$)} \\ 
    \hline
    \end{tabular}
}
\end{table}

\section{Conclusion} \label{sec:conclusion}

In this work, we used the Ordinal Pattern Transition Graph to classify transportation modes recorded as GPS trajectories. 
We transformed the GPS trajectory data, which is a time series, into Ordinal Patterns, and afterward, we transformed such patterns into the Transition Graph and the Probability Distribution of the pattern frequency. 
From the latter, we extracted two well-known Information Theory quantifiers, which are the Permutation Entropy and the Statistical Complexity; and from the former, we extracted a new feature, called the probability of self-transition, which is directly related to the temporal correlation of the original time series. 

The proposed feature presents better accuracy results than Permutation Entropy and Statistical Complexity, even when these two are combined. 
Hence we can affirm that the probability of self-transition satisfactorily characterizes the time series. Besides that, our feature has less dependence from the embedding dimension $D$ and embedding delay $\tau$, the needed parameters of Ordinal Pattern. 
Note that, although our proposal is validated here to the transportation mode classification, it may be generalized to time series in general, and can be used in other time series classification problems.

Furthermore, to the best of our knowledge, this is the first work that uses Information Theory quantifiers to transportation mode classification, showing that it is a feasible approach to this kind of problem.

For future work, we intend to study more about our proposed feature, especially its combination with other features, in order to achieve better time series characterization, and, consequently, better classification. 
Moreover, we intend to test our approach in different datasets to evaluate its robustness facing different problems.

\section*{Acknowledgement}

This work was partially funded by CNPq, FAPEMIG, and CAPES.

\bibliographystyle{unsrt}  
\bibliography{references}  

\end{document}